# Automatic Tuning of Interactive Perception Applications

**Qian Zhu**
The Ohio State University
zhuq@cse.ohio-state.edu

**Branislav Kveton**
Intel Labs Santa Clara
branislav.kveton@intel.com

**Lily Mummert**
Intel Labs Pittsburgh
lily.b.mummert@intel.com

**Padmanabhan Pillai**
Intel Labs Pittsburgh
padmanabhan.s.pillai@intel.com

## Abstract

Interactive applications incorporating high-data rate sensing and computer vision are becoming possible due to novel runtime systems and the use of parallel computation resources. To allow interactive use, such applications require careful tuning of multiple application parameters to meet required fidelity and latency bounds. This is a nontrivial task, often requiring expert knowledge, which becomes intractable as resources and application load characteristics change. This paper describes a method for automatic performance tuning that learns application characteristics and effects of tunable parameters online, and constructs models that are used to maximize fidelity for a given latency constraint. The paper shows that accurate latency models can be learned online, knowledge of application structure can be used to reduce the complexity of the learning task, and operating points can be found that achieve 90% of the optimal fidelity by exploring the parameter space only 3% of the time.

## 1 Introduction

Digital acquisition of video has become commonplace due to the availability of low-cost, digital cameras and recording hardware. Until recently, applications making use of video data have largely been limited to recording, compression, streaming, and playback for human consumption. Computer applications that can directly make use of video streams, for example as a medium for sensing the environment, detecting activities, or as a mode of input from human users, are now becoming areas of active research and development. In particular, a class of interactive perception applications that uses video and other high-data rate sensing for interactive gaming, natural gesture-based interfaces, and visually controlled robotic actuation is becoming increasingly important.

A great challenge in these applications is dealing with the tremendous computational costs involved, even for very rudimentary video analytics or computer vision algorithms, and the very low latencies needed for effective interaction, often limited to 50–100 ms. Current processors are not fast enough to deal with such applications. Two broad approaches are being used to achieve the required speeds. The first attempts to parallelize the execution of these applications on clusters of machines by transforming the applications into a pipeline or data flow graph of connected processing stages (Ramachandran et al., 2003; Allard et al., 2004; Pillai et al., 2009). The effectiveness of this approach depends greatly on the extent to which particular steps are parallelized. The second approach trades off result quality, or fidelity, with computation cost (Satyanarayanan and Narayanan, 2001). The algorithms used in interactive perception applications typically have many parameters that can have a significant effect on both fidelity and latency. For example, a feature extraction stage may have a tunable threshold of detection, a maximum iterations parameter, or a even switch to select from a set of alternative algorithms. The success of this approach depends on the tuning parameters available in the application. If dynamically adjustable, both degree of parallelism and algorithmic parameters provide an opportunity to control latency.

Given an application and a set of computing resources, which parameter settings yield the best combination of latency and fidelity? Application performance depends on a potentially large number of parameters that may have nonlinear effects and complex interactions. In addition, it also depends on the content of the input data and environmental conditions (e.g., processor configuration, external loads). Furthermore, dynamic parameter adjustments may require time to take effect, or have long settling times. The information needed to model application performance is unlikely to be known *a priori*, is difficult to reason about analytically, and is not static.

In this paper, we investigate a machine learning approach to modeling the effects of application tuning parameters on performance. First, we demonstrate how to predict the latency of workloads from streams of incoming data using state-of-the-art techniques for online convex programming (Zinkevich, 2003). We investigate both structured and unstructured approaches to the problem, and compare their accuracy and computational costs. Second, we build a system that learns how to predict the latency of workloads and also optimizes their fidelity. This problem is formulated as an online constrained optimization problem (Mannor and Tsitsiklis, 2006) with unknown latency constraints. We use $\varepsilon$-greedy strategies to explore the space of latency constraints, and show that an appropriate mixture of exploration and exploitation leads to a practical system for solving our problem. This is the first application of online learning with constraints to a structured real-world problem whose cost function is unknown.

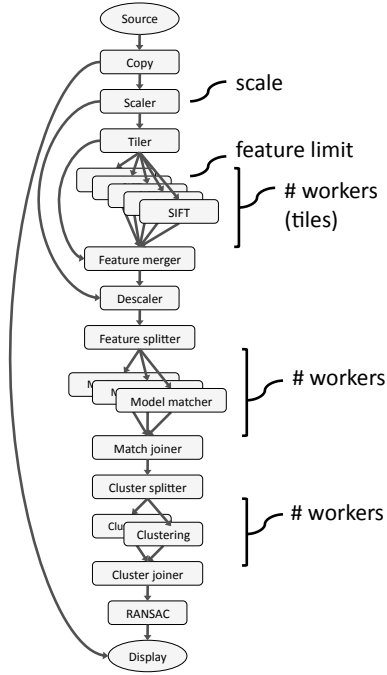

Figure 1: Data flow for pose detection application.

## 2 Modeling application performance

We consider parallel interactive perception applications that are structured as data flow graphs. The vertices of the graph are coarse-grained sequential processing steps called *stages*, and the edges are *connectors* which reflect data dependencies between stages. Stages interact only through connectors, and share no state otherwise. Source stages provide the input data to the application, for example as a stream of video from a camera. This data flows through and is transformed by multiple processing stages, which, for example, may implement a computer vision algorithm to detect when the user performs a particular gesture. Finally, the processed data is consumed by sink stages, which then control some actuator or display information to the user.

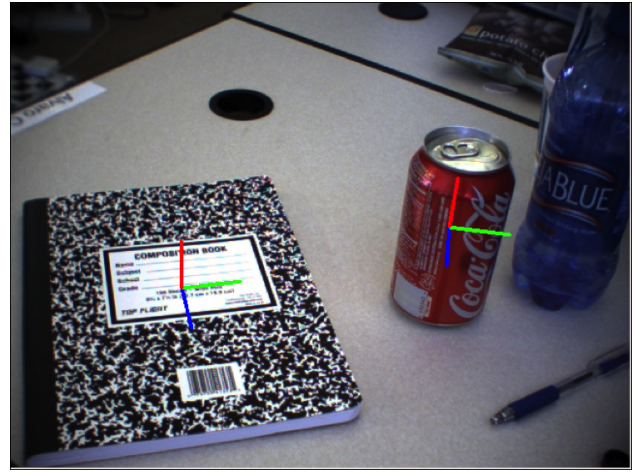

Figure 2: Pose detection output.

In the data flow model, concurrency is explicit – stages within an application can execute in parallel, constrained only by data dependencies and available processors. We use an application-independent runtime system (Pillai et al., 2009) to distribute and execute applications in parallel on a compute cluster. The system provides mechanisms to export and dynamically set tunable parameters, including both algorithmic parameters and controls for degree of parallelism (e.g., number of data-parallel operators). This system also monitors application performance, and provides interfaces for extracting latency data at the stage level.

### 2.1 Case studies

We study two applications in this paper. The first is an implementation of an algorithm for object instance recognition and pose registration used in robotics (Collet et al., 2009). As shown in the data flow of Figure 1, each image (frame from a single camera) first passes through a proportional down-scaler. SIFT features (Lowe, 2004) are then extracted from the image, and matched against a set of previously constructed 3D models for the objects of interest. The features for each object are then clustered by position to separate distinct instances. A random sample consensus (RANSAC) algorithm with a non-linear optimization is used to recognize each instance and estimate its 6D pose. Figure 2 shows the poses of two recognized objects. The implementation includes five tuning parameters: the degree of image scaling, a threshold on the number of features produced, and the degree of data parallelism used for feature extraction, model matching, and clustering. The details of such parameters are presented in Table 1. The listed default values of the parameters maximize application fidelity without regard to latency. To reduce the application latency, we can use a higher degree of data parallelism for each of those three stages. Extracting fewer features or scaling down the image will also accelerate the application. However, this leads to a degradation in application fidelity. As this application is intended for visual servoing of a robot arm, it requires very tight end-to-end latencies; our goal is a 50 ms latency bound.

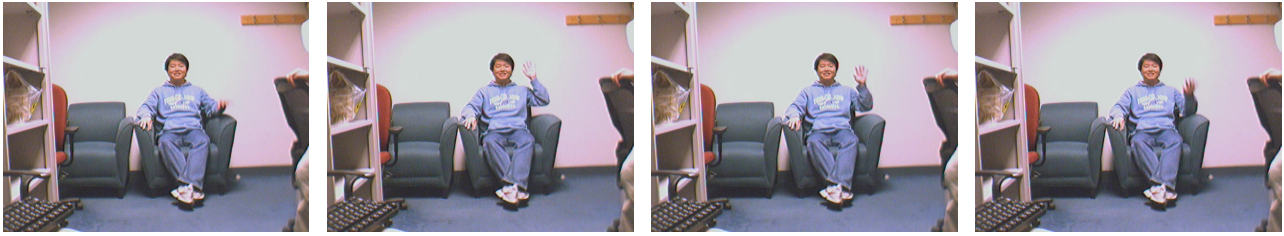

Figure 3: Viewer gesturing "channel up".

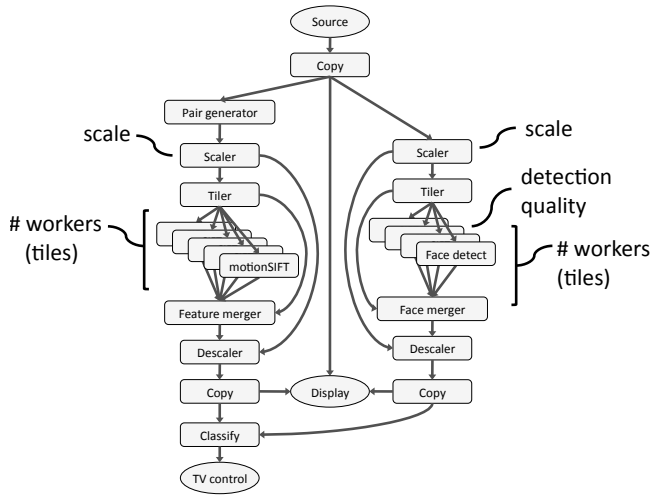

Figure 4: Data flow for gesture-based TV control.

The second application provides an interface to control a television via gestures (Chen et al., 2010). A camera positioned near the television observes a viewer as shown in Figure 3. Each video frame is sent to two separate tasks, face detection and motion extraction, shown in Figure 4. The latter accumulates frame pairs, and then extracts SIFT-like features that encode optical flow in addition to appearance. These features, filtered by the positions of detected faces, are aggregated over a window of frames using a previously-generated codebook to create a histogram of occurrence frequencies. The histogram is treated as an input vector to a set of support vector machines trained for the control gestures. The application exports five tuning parameters: the degree of scaling for each branch, the quality of face detection, and the degree of data parallelism for feature extraction and face detection. We describe the details of these parameters in Table 2. As above, the default values of the parameters are those that maximize application fidelity. Application latency can be reduced by invoking more instances of the features extraction and face detection stages. We can also scale down the frame size used by each branch to decrease the latency. However, using a higher degree of scaling or reducing the quality of face detection will degrade the application fidelity. For this application, low latency, on the order of 100 ms, is needed to achieve a responsive user interface. We note that in both of these applications, processing time of the vision algorithms are the primary contributers to latency, and dominate other sources, such as network transfer overheads.

### 2.2 Fidelity and latency models

Application fidelity varies as a function of the parameter settings. However, different parameters affect fidelity in different ways. For example, the degree of parallelism for a data parallel operation generally does not affect fidelity if the data items are independent. By contrast, a parameter that changes the content of the input data, such as by reducing the resolution of an image, can significantly affect fidelity. Fidelity is often difficult to quantify, especially when it relates to vague notions of perceptual quality. Even for specific measures like the accuracy of a detection algorithm, the true effect cannot be quantified without ground truth for a particular input. In the absence of a model based on ground truth, fidelity can be approximated relative to the "best" parameter settings, which are usually known.

Since stages can encompass arbitrary code, it is not practical to assume that application performance characteristics or the effects of tuning parameters on performance will be known *a priori* (e.g., via analytical modeling), particularly on varying hardware platforms. Nor is it practical to expect application programmers to supply even a subset of this information. The space of tuning parameters is large, and the settings may have non-linear effects and interactions. Finally, application performance may be data-dependent, and for this reason may change over time. Therefore, the application stages must be treated as a black box, with performance models learned online.

| **Variable** | **Type** | **Range** | **Default** | **Description** |
|---|---|---|---|---|
| $\mathcal{K}_1$ | continuous | $[1, 10]$ | 1 | The degree of image scaling |
| $\mathcal{K}_2$ | continuous | $[1, 2^{31}]$ | $2^{31}$ | A threshold on the number of produced features |
| $\mathcal{K}_3$ | discrete | $[1, 96]$ | 1 | The degree of data parallelism for feature extraction |
| $\mathcal{K}_4$ | discrete | $[1, 10]$ | 1 | The degree of data parallelism for model matching |
| $\mathcal{K}_5$ | discrete | $[1, 10]$ | 1 | The degree of data parallelism for clustering |

Table 1: Tuning parameters for pose detection application.

| **Variable** | **Type** | **Range** | **Default** | **Description** |
|---|---|---|---|---|
| $\mathcal{K}_1$ | continuous | $[1, 10]$ | 1 | The degree of image scaling for the left branch |
| $\mathcal{K}_2$ | continuous | $[1, 10]$ | 1 | The degree of image scaling for the right branch |
| $\mathcal{K}_3$ | discrete | $[0, 1]$ | 0 | The quality of face detection |
| $\mathcal{K}_4$ | discrete | $[1, 96]$ | 1 | The degree of data parallelism for feature extraction |
| $\mathcal{K}_5$ | discrete | $[1, 96]$ | 1 | The degree of data parallelism for face detection |

Table 2: Tuning parameters for gesture-based TV control application.

### 2.3 Reducing problem size

One can attempt to directly learn a model for the end-to-end application latency. However, because of the potentially large space of tuning parameters, and the need for fast and efficient learning, it is important to reduce the size of the learning task. Application structure, as given by the data flow graph, provides a natural way to partition the problem. A single stage may be affected by only a subset of the tuning parameters. For example, tuning parameters tend to influence localized sections of an application graph, and do not affect stages that are upstream or on parallel branches. In addition, some stages contribute little to total latency, or vary little, and may be modeled very simply (e.g., with an average). The end-to-end latency can then be obtained by combining the predictions of the stage models according to the critical path through the data flow graph. For example, in Figure 4, the end-to-end latency is sum of the latencies of the source, copy, classify, and sink stages, and the maximum of the latencies of the face detection and motion extraction subgraphs, computed similarly. The total complexity of learning all of the individual stage models with a few parameters for each can be significantly lower than learning a single model based on all parameters.

We reduce the automatic tuning problem as follows. We first use a few observations of stage latencies to identify a set of *critical stages*, based on their contribution to end-to-end latency. A dependency analysis is performed to identify the parameters that affect each critical stage. Specifically, a parameter is associated with a critical stage if the correlation between the value of the parameter and the stage latency exceeds a threshold (0.9 in this work). Then, with additional periodic observations, we explore the parameter space and learn a predictor as a function of the relevant tuning parameters for each critical stage. Non-critical stages are modeled with a moving average. End-to-end latency is predicted using the stage models by computing the critical path through the data flow graph. A solver is used to search for operating points that maximize fidelity for a given latency constraint. Changes in parameter settings are then applied to the running application.

## 3 Problem formulation

An application is defined as a tuple $(G, \mathcal{K}, L)$, where $G$ is a data flow graph (Section 2), $\mathcal{K} = \mathcal{K}_1 \times \cdots \times \mathcal{K}_m$ is a space of dynamically tunable parameters $\mathcal{K}_1, \ldots, \mathcal{K}_m$, and $L$ is a latency bound. Specifically, $G = (V, E)$ is a directed graph consisting of $n$ computation stages $V = \{v_1, \ldots, v_n\}$ and connectors $E = \{e_{ij} |$ stage $j$ requires data from stage $i\}$. Nodes of the graph $G$ are weighted by computation latency. The weight $w_i$ of node $v_i$ is the latency of a single execution of stage $i$. The latency of the application is the length of the critical path $C$ through $G$, which is given by $c = \sum_{i: i \in C} w_i$. For simplicity, we omit inter-stage communication latency from this formulation, which can be incorporated by adding edge weights that represent communication costs.

### 3.1 Online learning with constraints

Let $\mathbf{k}_t = (k_{t1}, \ldots, k_{tm})$ be values of the tunable parameters $\mathcal{K}_1, \ldots, \mathcal{K}_m$ at time $t$, $\mathbf{x}_t \in \mathcal{X}$ be the data gathered at time $t$; and $r(\mathbf{x}_t, \mathbf{k}_t)$ and $c(\mathbf{x}_t, \mathbf{k}_t)$ be the corresponding fidelity and latency, respectively. Then the problem of maximizing fidelity subject to latency constraints can be formulated as:

$$\max_{\mathbf{k}_1, \ldots, \mathbf{k}_T} \sum_{t=1}^{T} r(\mathbf{x}_t, \mathbf{k}_t) \quad (1)$$
$$\text{s.t.} \quad c(\mathbf{x}_t, \mathbf{k}_t) \leq L.$$

The challenge in solving our problem online is that neither $c$ nor $r$ are usually known in advance. Under these assump-tions, it is hard to solve the problem online (Mannor and Tsitsiklis, 2006). In fact, the optimal offline solution may not be attainable in the online setting.

If both $c$ and $r$ are known, the optimal action at time $t$ is:

$$\mathbf{k}_t^* = \arg\max_{\mathbf{k}} r(\mathbf{x}_t, \mathbf{k}) \mathbb{1}\{c(\mathbf{x}_t, \mathbf{k}) \leq L\}. \quad (2)$$

In the rest of the paper, we assume that the reward function $r$ is known and focus on learning of the cost function $c$. The learning step is alternated with solving Equation 2 using an $\varepsilon$-greedy policy (Sutton and Barto, 1998), and the result is a practical solution to our problem. How to choose $\varepsilon$ is discussed in Section 4.

For simplicity, we assume that the cost function $c$ does not change in time. If $c$ is represented in a tabular form, then we can learn an $\varepsilon$-close approximation to $c$ with probability $1 - \delta$ in polynomial time in $|\mathcal{K}|$ and $|\mathcal{X}|$ (Brafman and Tennenholtz, 2003). Unfortunately, learning this representation may not be practical or even possible. For instance, if all $\mathcal{K}_1, \ldots, \mathcal{K}_m$ are discrete, then $|\mathcal{K}|$ is exponential in $m$. Note that each of the problems in Figures 1 and 4 have 5 tunable parameters, and some of the parameters are continuous (Tables 1 and 2).

### 3.2 Learning of the cost function

To take the structure of our problem into account and speed up learning, we learn a regressor $f$ of the cost function. The regression problem can be formulated as (Smola and Schölkopf, 2004):

$$\min_{f \in \mathcal{H}_K} \frac{1}{T} \sum_{t=1}^{T} V_\varepsilon(f, (\mathbf{x}_t, \mathbf{k}_t), c(\mathbf{x}_t, \mathbf{k}_t)) + \gamma \|f\|_K^2, \quad (3)$$

where:

$$V_\varepsilon(f, (\mathbf{x}, \mathbf{k}), y) = \max\{|f(\mathbf{x}, \mathbf{k}) - y| - \varepsilon, 0\} \quad (4)$$

denotes the $\varepsilon$-*insensitive loss*, $c(\mathbf{x}_t, \mathbf{k}_t)$ is the cost at time $t$, $f$ is a function from some *reproducing kernel Hilbert space (RKHS)* $\mathcal{H}_K$, and $\|\cdot\|_K$ is the norm that measures the complexity of $f$. The tradeoff between the minimization of the losses $V_\varepsilon(f, (\mathbf{x}_t, \mathbf{k}_t), c(\mathbf{x}_t, \mathbf{k}_t))$ and the regularization of $f$ is controlled by the parameter $\gamma$. In all of our experiments, $\gamma = 0.01$.

Unfortunately, the optimal solution to our regression problem minimizes the expected error $\mathbb{E}_{\mathbf{x},\mathbf{k}}[|f(\mathbf{x},\mathbf{k}) - c(\mathbf{x},\mathbf{k})|]$ rather than the max-norm error $\|f(\cdot,\cdot) - c(\cdot,\cdot)\|_\infty$, which would be more suitable for our task. On the other hand, note that the two errors can be related though the Lipschitz factors of $c$ and $f$. In particular, if both $c$ and $f$ are smooth, the minimization of the average error leads to minimizing the max-norm error. These trends are shown in Section 4.

### 3.3 Online learning of the cost function

Online learning of the regressor (3) can be formulated as an online convex programming problem. Online convex programming (Zinkevich, 2003) involves a convex feasible set $\mathcal{F} \subset \mathbb{R}^n$ and a sequence of convex functions $\ell_t : \mathcal{F} \to \mathbb{R}$. At each time step $t$, a decision maker chooses some action $f_t \in \mathcal{F}$ based on the past functions $\ell_1, \ldots, \ell_{t-1}$ and actions $f_1, \ldots, f_{t-1}$. The goal is to minimize the regret:

$$\sum_{t=1}^{T} \ell_t(f_t) - \min_{f \in \mathcal{F}} \sum_{t=1}^{T} \ell_t(f). \quad (5)$$

Intuitively, we want to choose online a sequence of actions $f_1, \ldots, f_T$ such that their costs are close to the optimal action $\arg\min_{f \in \mathcal{F}} \sum_{t=1}^{T} \ell_t(f)$ in hindsight.

The above regret is minimized on the order of $O(\sqrt{T})$ by the gradient update:

$$f_{t+1} = P(f_t + \eta \nabla \ell_t(f_t)), \quad (6)$$

where $\eta = \sqrt{T}$ is a learning rate, $\nabla \ell_t(f_t)$ is the gradient of the function $\ell_t$ at $f_t$, and $P(\cdot)$ is a projection to the feasible set $\mathcal{F}$. To minimize the SVM regression objective (3), the cost function $\ell_t$ is chosen as:

$$\ell_t(f) = V_\varepsilon(f, (\mathbf{x}_t, \mathbf{k}_t), c(\mathbf{x}_t, \mathbf{k}_t)) + \gamma \|f\|_K^2. \quad (7)$$

For linear SVMs, the cost function further simplifies as:

$$\ell_t(f) = V_\varepsilon(f, (\mathbf{x}_t, \mathbf{k}_t), c(\mathbf{x}_t, \mathbf{k}_t)) + \gamma \|f\|_2^2. \quad (8)$$

Online learning of non-linear regressors can be done in two ways. First, we can directly solve the kernelized version of the problem (Kivinen, Smola, and Williamson, 2004). The main problem with this approach is that we have to maintain a compact representation of the kernel matrix. The second option is to expand the original feature space by non-linear features and learn a linear regressor in the new space. This technique is suitable for quadratic and cubic kernels, for instance. In the experimental section, we use cubic kernels and adopt the latter approach.

Unfortunately, the cubic expansion of the feature space can be costly. To make our approach applicable to larger problems, we take advantage of how the critical path $c$ of a data flow graph decomposes based on the structure of the graph (Section 2.3). For instance, note that the cost regressor for the problem in Figure 4 can be rewritten as:

$$f(\mathbf{x}_t, \mathbf{k}_t) = \max\{f_L(\mathbf{x}_t, \mathbf{k}_t), f_R(\mathbf{x}_t, \mathbf{k}_t)\}, \quad (9)$$

where $f_L(\mathbf{x}_t, \mathbf{k}_t)$ and $f_R(\mathbf{x}_t, \mathbf{k}_t)$ correspond to the left and right subgraphs, and are defined on the subspaces of $\mathcal{X}$ and $\mathcal{K}$. To solve this problem in a structured manner, we learn a regressor for each of the subspaces, and then combine them by a deterministic function, such as max for parallel stages, and sum for sequential structures.

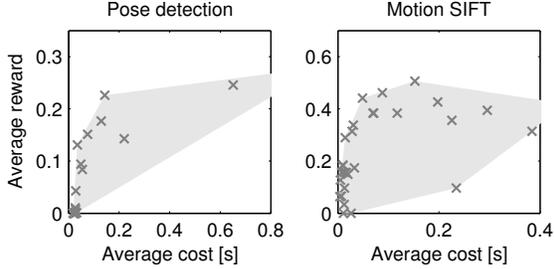

Figure 5: Average rewards and costs of 30 action configurations from our action space (gray crosses). The convex hull represents payoffs, which are feasible by playing a randomized strategy over the 30 action configuration.

## 4 Evaluation

In this section, we evaluate our approach in three ways. First, we compare the accuracy of latency predictors learned online with those learned offline. Second, we study the effect of using application structure by comparing the accuracy of a single predictor for the application with predictors learned for individual stages and combined as described in Section 3.3. Finally, we examine the tradeoffs between exploration (learning) and exploitation (optimization) to determine high-fidelity operating points for a given latency bound.

### 4.1 Methodology

We conducted our experiments using the system and applications described in Section 2. For each application we used as input a video sequence annotated with ground truth. For the pose detection application, the video consists of a series of objects in different positions and orientations, and the ground truth is the object label with its measured translation and rotation in each frame. For the gesture-based TV control application, referred to henceforth as "Motion SIFT" after the low-level feature it uses to represent motion, the video consists of a single user performing control gestures (as shown in Figure 3), and the ground truth is the label of the gesture occurring in the frame, if any.

For greater experimental control and the repeatability of results, our experiments are done on a set of execution traces. These were collected on a cluster of 15 servers connected with a 1 Gbps Ethernet switch. Each server has two Intel® 2.83 GHz Xeon® E5440 processors (8 cores total) and 8 GB of memory, and runs Ubuntu Linux 8.04. For each application, we created 30 configurations by selecting random valid values for the tunable parameters in Figures 1 and 4. We ran each of these static configurations on a sequence of 1000 frames, collected performance logs from the runtime, and extracted latency measures for each frame. We use the set of configurations as a point-based approximation of the total space, and use the traces as predefined alternative futures between which the simulated system switches as our algorithm executes.

To quantify fidelity, for each application, we used ground truth to define the function $r$ from Equation 1. For pose detection, $r$ measures recognition correctness and pose accuracy:

$$r(\mathbf{x}_t, \mathbf{k}_t) = \frac{1}{n} \sum_{i=1}^{n} \mathcal{R}_{it} e^{-(w_\tau \tau_{it} + w_\theta \theta_{it})} \quad (10)$$

where $n$ is the number of objects in the scene, $\mathcal{R} = \{0, 1\}$ indicates whether an object is recognized, $\tau$ is the translation error, and $\theta$ is the rotation error. Weights $w_\tau$ and $w_\theta$ were set to $0.7$ and $0.3$, respectively. For Motion SIFT, $r$ is the $F_1$-measure for classification performance:

$$r(\mathbf{x}_t, \mathbf{k}_t) = 2 \frac{P_t R_t}{P_t + R_t} \quad (11)$$

where $P$ is precision and $R$ is recall. Figure 5 shows the average latency (cost) and average fidelity (reward) for each application and configuration (action).

### 4.2 Predicting latency

In the first experiment (Figure 6), we study how the quality of latency predictors depends on their complexity. The predictors are learned online using linear, quadratic, and cubic kernels. In particular, at each time step, we randomly sample an action and then update the predictors as described in Section 3.3.

Figure 6 shows that the errors of our predictors tend to decrease over time. The increase in the pose detection dataset at frame 600 corresponds to a change in the scene, in which a notebook (Figure 2) appeared. This increased the number of SIFT features in the scene and consequently the computational requirements to process a single frame. In general, cubic predictors yield the smallest errors, and all predictors are almost as good as their offline counterparts. In addition, note that the max-norm errors of our latency predictors also decrease over time. This observation is consistent with our expectation that the minimization of the average error leads to minimizing the max-norm error if both the cost function and its regressor are smooth (Section 3.2).

### 4.3 Effect of structure

In the second experiment (Figure 7), we compare structured and unstructured approaches to learning latency predictors. The predictors are cubic and learned online as described in Section 4.2.

Figure 7 shows that the expected errors of unstructured and structured latency predictors are almost identical. The main advantage in using the structured predictors is that their cubic feature space expansion is smaller. For instance, on the motion SIFT dataset, it takes 30 and 56 features to describe the structured and unstructured spaces, respectively. Thus, updating of the structured predictor should be twice as fast in practice. If our problems involved hundreds of variables, the speedup would likely be much more significant.

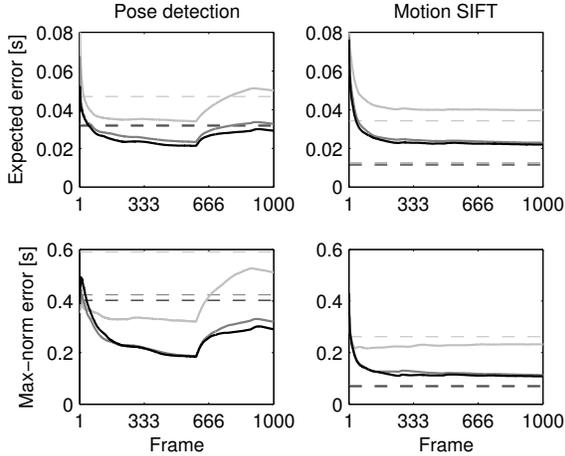

Figure 6: Comparison of linear (light gray lines), quadratic (dark gray lines), and cubic (black lines) latency predictors. The predictors are learned online and compared by the cumulative average of their expected and max-norm errors up to each frame. The errors of the corresponding offline predictors are shown as dashed lines.

Finally, note that the max-norm errors of structured latency predictors can be significantly smaller than the errors of the unstructured predictors. One way of explaining this results is that the expected and max-norm errors are coupled more tightly on the subspaces of the problems. In turn, the minimization of the expected error on the subspaces results in a smaller total max-norm error.

### 4.4 Online learning with constraints

In the last experiment (Figure 8), we use our latency predictors to build a control policy that maximizes fidelity subject to latency constraints. The resulting controller is simply an $\varepsilon$-greedy policy, where $\varepsilon$ defines the amount of exploration. To find the optimal exploration rate, we vary $\varepsilon$ and measure the corresponding average fidelity and constraint violation, which is given by $\mathbb{E}_{\mathbf{x}_t, \mathbf{k}_t}[\max\{c(\mathbf{x}_t, \mathbf{k}_t) - L, 0\}]$. Our results are compared to the payoff of randomized strategies in our action space (Figure 5). Intuitively, we want to achieve close-to-zero constraint violation and maximize the fidelity at this operating point.

Figure 8 illustrates the performance of our policies for various exploration rates $\varepsilon$ and latency bounds $L$. As $\varepsilon$ varies, the policies usually follow a U-shaped curve. In particular, when $\varepsilon$ is too small, the estimate of the latency function is uncertain, which results in policies that significantly violate the latency bound $L$. On the other hand, when $\varepsilon$ is close to one, we mostly explore and perform poorly with respect to the main objective, which is the maximization of fidelity.

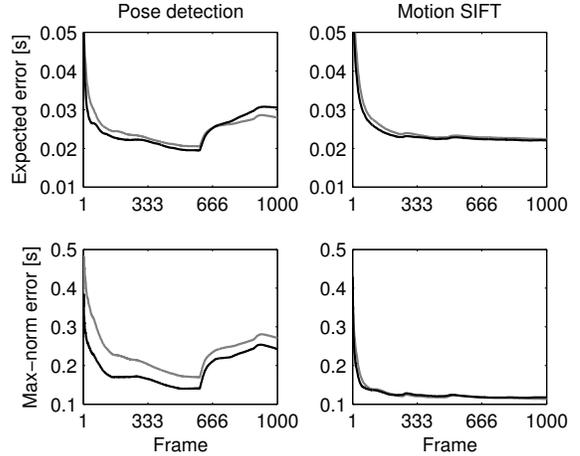

Figure 7: Comparison of unstructured (dark gray lines) and structured (black lines) latency predictors. The predictors are learned online and compared by the cumulative average of their expected and max-norm errors up to each frame.

An ideal exploration rate $\varepsilon$ guarantees that we both explore enough to learn the cost function and exploit enough to optimize our main objective. One of the reasonable choices is $\varepsilon = 1/\sqrt{T}$, which yields $\varepsilon = 0.03$ when $T = 1,000$. For this setting, the regret (5) is sublinear in $T$, and the proportion of exploitation to exploration $(T - \sqrt{T})/\sqrt{T}$ increases polynomially with $T$. Thus, the performance of our system should improve over time.

The diamonds in Figure 8 mark the operating points of the $(1/\sqrt{T})$-greedy policies. In all experiments, these policies yield high rewards and low constraint violations. In particular, note that our rewards are always within 90 percent of the optimum. Moreover, the average constraint violation in all experiments is about 0.03 second and never exceeds 0.1 second. When measured relatively to the latency bound $L$, the average and worst-case constraint violations are 23 and 50 percent, respectively.

## 5 Related work

Machine learning has been used in several contexts to predict and tune interactive and parallel applications. In the domain of mobile devices, Narayanan and Satyanarayanan (Narayanan and Satyanarayanan, 2003) have modeled the latency and battery consumption of interactive mobile applications as a function of tunable fidelity settings. Initially, random sampling is used offline to train linear predictors on user-supplied terms, followed by online refinement of model coefficients. In contrast, our approach automatically learns performance and fidelity relationships online.

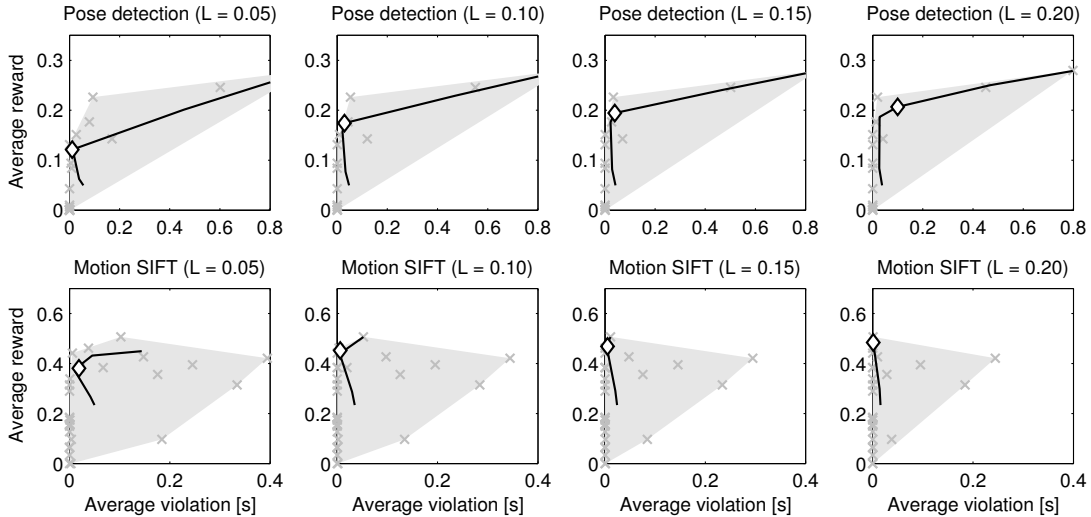

Figure 8: Average rewards and constraint violations of our polices (black lines) for various exploration rates $\varepsilon$. The policies are compared with respect to the payoff of randomized strategies over our action space (gray regions). The operating points $\varepsilon = 1/\sqrt{T}$ are marked by diamonds.

For scientific computing workflows, the NIMO system (Shivam, Babu, and Chase, 2006) employs active learning to estimate completion times of applications on particular datasets with heterogeneous processing and storage resources. It employs multivariate linear regression and a design of experiments approach to exploring tradeoffs and interactions between attributes, since each experiment can take hours or days to complete. In our system, tunable parameters can be changed dynamically during execution, facilitating rapid online exploration, although this must be balanced with the need to avoid large perturbations to the system. Also in this domain, Artificial Neural Networks (ANNs) have been used to construct performance models of partitioned parallel worksets (Li et al., 2009), and have been shown as effective as polynomial regression approaches in modeling scientific computing applications (Lee et al., 2007), but with less domain-specific knowledge required for model formulation. Active learning was used to reduce ANN-based model error by selecting samples that generated the highest CoV from an model ensemble (Singh et al., 2007). Finally, reinforcement learning was used to select a representation format for sparse matrices based on the characteristics of the data (e.g., number of nonzero elements) to minimize execution time for a fixed set of multiplications (Armstrong and Rendell, 2008).

For web applications, Xi *et al.* (Xi et al., 2004) model server performance as a black-box function of parameters, and employ a smart hill climbing algorithm with weighted Latin hypercube sampling to find high performance configurations offline. This method requires advance planning or careful tracking of all prior samples. Reinforcement learning has been used for online tuning of configuration parameters for a multi-tiered web application (Chen et al., 2009) to optimize a combination of throughput and response time.

In contrast to these, our work focuses on the tuning of interactive parallel applications, with a focus on fidelity and latency. We use an online learning algorithm (Zinkevich, 2003) with strong theoretical guarantees to learn the latency function, and take advantage of application structure to reduce the complexity of the learning task.

## 6 Conclusion

This paper has shown a practical application of online learning to estimate performance and dynamically adjust tunable parameters in real-world interactive perception applications. The approach outlined can readily model latency characteristics of a complex application. It uses application structure to both reduce the learning task complexity and improve expected and max-norm prediction error. Finally, it trades off exploration and exploitation to find high-fidelity operating points for a given latency bound. This work enables a critical new capability in the emerging class of interactive perception applications: the ability to automatically adapt to the particulars of a distributed set of parallel computing resources and dynamically changing workload characteristics. This work can be seen as a general template for practical application of machine learning techniques to important real-world problems. Among the areas of future work, we plan to incorporate models for network latency, and study exploration strategies that take into account the cost of changing parameter settings and the expected improvement in fidelity.

**Acknowledgements**

We wish to thank Ling Huang for helpful discussions that led to the improvement of this paper, and Alvaro Collet and Robert Chen for their assistance in obtaining annotated data sets.